# Predicting the Effects of News Sentiments on the Stock Market

Dev Shah
School of Computing
Queens University
Kingston, ON, Canada
16ds57@cs.queensu.ca

Haruna Isah
School of Computing
Queens University
Kingston, ON, Canada
isah@cs.queensu.ca

Farhana Zulkernine
School of Computing
Queens University
Kingston, ON, Canada
farhana@cs.queensu.ca

*Abstract*— Stock market forecasting is very important in the planning of business activities. Stock price prediction has attracted many researchers in multiple disciplines including computer science, statistics, economics, finance, and operations research. Recent studies have shown that the vast amount of online information in the public domain such as Wikipedia usage pattern, news stories from the mainstream media, and social media discussions can have an observable effect on investors' opinions towards financial markets. The reliability of the computational models on stock market prediction is important as it is very sensitive to the economy and can directly lead to financial loss. In this paper, we retrieved, extracted, and analyzed the effects of news sentiments on the stock market. Our main contributions include the development of a sentiment analysis dictionary for the financial sector, the development of a dictionary-based sentiment analysis model, and the evaluation of the model for gauging the effects of news sentiments on stocks for the pharmaceutical market. Using only news sentiments, we achieved a directional accuracy of 70.59% in predicting the trends in short-term stock price movement.

## I. INTRODUCTION

The decision of when to buy or sell shares is an interesting research challenge in the stock market. Such decisions are being negotiated daily in the stock market across the globe. Indices were created as a way to measure the relative value of stocks in a determined group in the market [1]. The idea about whether stock markets can be predicted has kept economists and investors very busy for decades. The two distinct trading philosophies for stock market prediction are fundamental and technical analysis [2]. While technical analysis focuses on the study of market actions through the use of charts, fundamental analysis concentrates on the economic forces of supply and demand that cause the stock price to move higher, lower, or remain the same [3]. Stock price prediction has attracted many researchers in multiple disciplines such as computer science, statistics, economics, finance, and operations research [4]. These research efforts have so far produced several methods for forecasting the future direction of the stock market [5].

In the discipline of computer science and the field of machine learning, some of the computational methods that have been utilized for stock prediction include Bayesian Networks [1], Artificial Neural Networks [6], and Support Vector Machines [7]. According to Guresen et al, the common indicators or variables that have been utilized for traditional stock price prediction include time series data from the literature, banking sector data series, daily closing values of stock market indices, and monthly weighted stock indices [8]. Recent studies, however, has shown that the vast amount of online information such as Wikipedia usage pattern and news stories from mainstream and social media sources can have an observable effect on investor's opinions towards financial markets [9]. Researchers have, therefore, begun to incorporate Natural Language Processing (NLP) methods in the process of stock price prediction [4].

Sentiment analysis, the task of mining subjective feelings expressed in the text [10], has been found to play a significant role in many applications such as product recommendations, healthcare, politics, and in surveillance [11]. The expression of moods and emotions in a large amount of social media data is very important in gauging the opinions of investors [2]. Twitter data has recently gained traction in predicting stock prices based on public sentiments [12]. This is because Twitter data is already public and thus automatically and very quickly influences stock prices. However, quality measures must be put in place in order to use Twitter data because of its widespread use by malicious users to promote or devalue products, services, ideas, and ideologies [13]. The reliability of the models for stock market prediction is important as it can directly affect economy and lead to financial loss. News data are sources of information that can be relatively relied upon. In this paper, we retrieved, extracted, and analyzed the effects of news sentiments on stocks in the pharmaceutical sector since it is more sensitive to news and comments in the media.

The rest of the paper is organized as follows. Section II presents a background on stock market prediction and sentiment analysis. A dictionary-based news sentiment analysis model is presented in Section III. The evaluation of the model is presented in section IV. Section V concludes the paper with a discussion of the future work.

## II. BACKGROUND

Stock price prediction is very important in the planning of business activities. There are many microeconomic and macroeconomic variables that can lead to the movement of price in the stock market [2]. Traditional technical analysts have developed many indices and sequential analytical methods that may reflect the trends in the movements of the stock price [4]. However, in addition to historical prices, the societal mood is seen to be playing a significant role in stock price movement. The overall social mood with respect to a given company is now being considered as an important variable which affects the stock price of that company. Online information such as public opinions [2], social media discussions [12], and Wikipedia activities [9] are being used to study their impact on pricing and investors' behaviors taking into account the risk factors due to biased and malicious posts. Researchers have shown that integrating models based on historical stock prices with the data from the mainstream and social media platforms can improve the predictive ability of the analytical models of stock price prediction [2][4].

Sentiment analysis is an NLP technique for mining and evaluating public opinions expressed in text/speech [10]. The increasing interest in sentiment analysis is partly due to its potential applications [11]. Major elements of sentiment analysis methods include the sentiment (opinion expressed), target object or topic, time of expression, person expressing the opinion, the reason for the expression, the opinion qualifier, and the opinion types. Sentiment analysis is divided into several levels according to different discourse granularity, such as document, sentence, entity, and aspect. It can also be implemented in a supervised, unsupervised, or semi-supervised approach [10]. This paper adopts a dictionary-based learning approach at the entity level to analyze the effects of sentiments expressed in news articles on stocks.

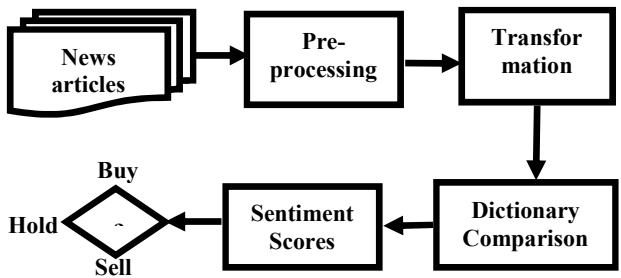

Fig. 1. A dictionary-based news sentiment model.

III. A DICTIONARY-BASED SENTIMENT ANALYSIS MODEL

A general approach to understanding the sentiment behind news articles is to identify important words and their polarity in the news articles. Dictionary or lexicon-based approaches are robust as they provide good cross-domain performance. A dictionary-based news sentiment analysis model is shown in Fig. 1. First, a reliable source of news data was selected, and a web scraper was developed to collect the data. Next a pre-processing step was executed to clean and tokenize the data. The pre-processed data is then transformed to a suitable format (such as n-grams). The transformed data is then compared against words in a predefined dictionary, which consisted of domain specific words and phrases and their associated polarity strength in terms of sentiment. The output of the comparison step is an aggregated list of news articles with their corresponding sentiment scores. These scores will then be analyzed and cross-validated against the stock prices to understand their effect. The overall goal of the model is to enhance the stock trading decision-making process. Based on the score, the decision may be to 'buy', 'sell' or 'hold' which means do nothing. For instance, if the overall score of a news is positive and is above a predefined threshold then the stock will be bought and vice versa.

*A. Implementation*

In this study, we used moneycontrol.com[1], one of the topmost and reliable stock-specific news providers in India as the source of news articles while Investing.com[2] was used as the source for retrieving detailed intraday (30-minute gap) pricing data for each stock.

*1) Data Sources and Preprocessing*

A dataset consisting of specific news articles from the past six months, pertaining to stocks which form the Nifty Pharma Index was chosen to perform the sentiment analysis. First, a web scraper was developed to handle the format of the article links that needed to be scrapped. Next, the news articles were fetched using the moneycontrol.com news API. The entire text of the articles was fetched from the specific URL's using Beautiful Soup library in Python. The scraper was designed in such a way that only articles from the previous six months were stored for each company. Indian pharmaceutical companies are highly dependent on the United States Food and Drug Administration's (FDA) inspections and drug approvals because the majority of their revenues come from export. Therefore, articles containing the words 'US', 'USA', 'USFDA' in their headline were chosen for further analysis. Also, headlines having keywords like 'Q1', 'Q2' etc. were chosen to analyze the impact of quarterly results on the respective stocks. Since there are high chances of finding duplicate news articles, only unique news articles were selected to form the news corpus. The text was then preprocessed to handle punctuation, special characters and white spaces by simply removing them from the text.

*2) Data Transformation and Model Configuration*

A Python library called "pattern" was used for transforming the text corpus into numerical vectors. For each news article, the main text was extracted and transformed into n-grams. Initially unigrams (where the number of word tokens is one) were used for the sentiment analysis. However, the unigrams did not capture the context of the text. For instance, the word 'decline' is generally a negative word, however, if the phrase is 'costs declined' then it is positive for the company. Eventually, bigrams (where the number of word tokens is two) and trigrams (where the number of word tokens is three) were created from the text corpus. These helped to understand the polarity of individual words and the surrounding phrases as well, which allowed capturing the context in a much better way. Stemming i.e., reducing words to their base forms such as 'declining' to 'decline' is preferred for a phrase and was applied to processing bigrams and trigrams. Fig. 2 and Fig. 3 are examples of n-grams formed for the news articles of Alkem laboratories (NSE:ALKEM) stock.

---

[1] http://www.moneycontrol.com

[2] https://www.investing.com/

Fig. 2. Unigrams from a news article.

Fig. 3. Bigrams from a news article.

Due to the unavailability of dictionaries specific to the financial markets, we manually created a dictionary that contains words and phrases which carried specific meaning pertaining to pharmaceutical companies. The dictionary was created by leveraging author's domain expertise and thorough analysis of news articles over the years. The dictionary consisted of 100 words which cover a broad spectrum of topics specific to the pharmaceutical sector. Each entry in the dictionary was tagged with positive, negative, or neutral class of sentiment. The generated n-grams were then compared against the dictionary, and if a match was found then each matching word was assigned a positive or negative polarity. The positive and negative words were then counted and the sentiment scores were generated based on their frequency. For instance, if there are three positive words then the sentiment would be +3. Words with neutral sentiment do not affect the score. For all the news articles, the scores were cross-validated against the stock prices to understand their effects. A straightforward portfolio strategy was devised to gauge the performance of the sentiment analysis model. Based on the score, the decision would be to 'buy', 'sell' or 'hold' (do nothing). If the overall score of a news is positive and above a predefined threshold then the stock will be bought after the news is out and vice versa.

IV. EVALUATION

We evaluate our model using a few selected pharma stocks. Several assumptions were made while constructing this portfolio strategy. First, the stock was assumed to be bought or sold within thirty minutes of the release of a news. Second, the returns were calculated based on the buy or sell price minus the closing price for that day. Furthermore, if the model suggests a buy decision based on the sentiment score, then it is considered a correct prediction only if the stock gains more than 0.5%. Hence a threshold of 0.5% was implemented for 'buy' and 'sell' decisions and 1% for 'hold' decisions. For example, if the model suggests to 'hold' a stock then it is essentially saying that the stock will not move either way for more than 1%, if it does, then the prediction is incorrect. Our sentiment analysis model achieved a 70.59% accuracy in predicting daily stock movement direction. While evaluating model's performance, we observed that there weren't any significant changes in the base index Nifty Pharma, thereby eliminating the possibility of broader market forces influencing the prices of constituent stocks.

Fig. 4, 5, and 6 show the results of the buy, sell and hold decisions made by the sentiment analysis model based on the scores for different stocks. Green arrow stands for buy decision or up move, red arrow stands for sell i.e. down move, and yellow arrow stands for hold decision i.e. no move. The red arrow in Fig. 4 depicts the rightly predicted sharp fall in NSE:ALKEM, when the USFDA issues 13 observations on its plants, which is indeed a negative news for the stock.

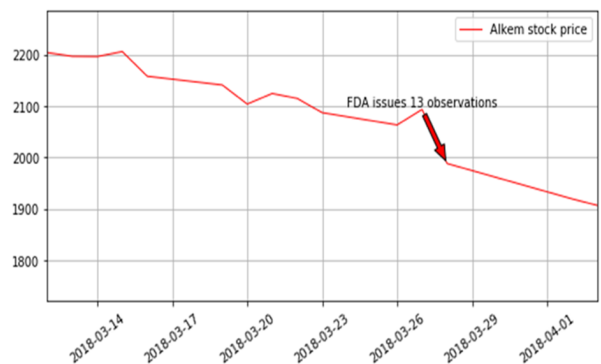

Fig. 4. Sell decisions on NSE:ALKEM.

Fig. 5 depicts the correct 'buy' predictions made by the model on NSE:LUPIN based on the different news published over a three month period. Fig. 6 gives an example of how the model predicts buy, sell and hold decisions for NSE:SUNPHARMA based on the news data. For instance, the yellow arrow indicates a hold decision made by the model based on a news article which suggests that the company posted disappointing numbers for Q2 (second quarter).

However, the article mentions that it is inline with the market expectations and hence there is no fall in the stock. Thus, the model is able to rightly gauge the sentiment as 'neutral'.

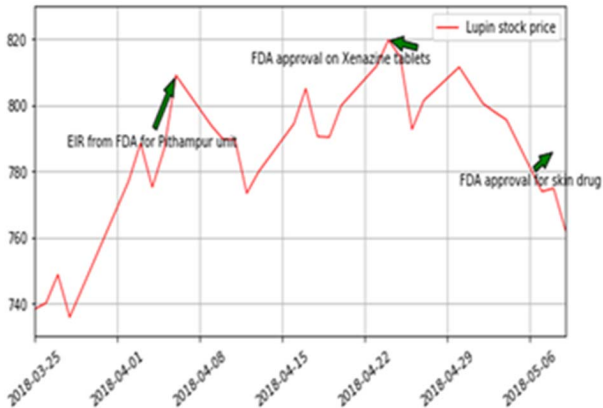

Fig. 5. Buy decisions on NSE:LUPIN.

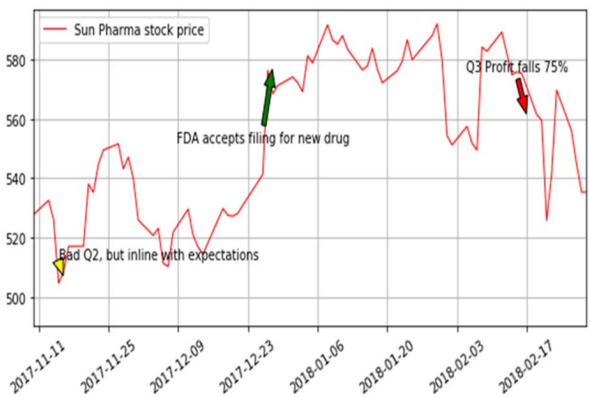

Fig. 6. Buy, sell and hold decisions on NSE:SUNPHARMA

## V. CONCLUSION

This study is a niche application of sentiment analysis in gauging the effects of news sentiments on stocks for the pharmaceutical market sector. One major contribution of this work is a sentiment analysis dictionary [3]. The sentiment scores obtained from the analysis of the news articles is a powerful indicator of stock movements and can be used to effectively leverage the prediction of short-term trends. We believe that the reason the model is able to achieve an accuracy of 70.59% with the dictionary-based approach is that the dictionary was specifically created for this particular sector by researching and leveraging domain expertise.

In terms of future work, a word weighting approach can be implemented in the sentiment analysis pipeline to determine which word is more important and then assign the sentiment score accordingly. Since domain specific sentiment lexicon cannot be easily scaled manually, expansion of the dictionary leveraging well established lexicons [14] would make our model scalable for other sectors as well. Lastly, creating a hybrid model using traditional statistical classification models such as Auto Regressive Integrated Moving Average (ARIMA) and machine learning models such as the long short-term memory (LSTM) neural network model, using stock prices and sentiment analysis model based on social media and news data, may provide more accurate predictions of long and short-term stock price movements.

---

[3] https://github.com/queensbamlab/NewsSentiments/blob/master/dict.csv